\documentclass[sigconf]{acmart}

\setcopyright{none}
\renewcommand\footnotetextcopyrightpermission[1]{}

\AtBeginDocument{%
  }

\begin{document}

\title{Agent Safety Is Action Alignment}

\author{Shawn Li}
\email{li.li02@usc.edu}
\affiliation{%
  \institution{University of Southern California}
  \city{Los Angeles}
  \state{California}
  \country{USA}
}

\author{Yue Zhao}
\email{yue.z@usc.edu}
\affiliation{%
  \institution{University of Southern California}
  \city{Los Angeles}
  \state{California}
  \country{USA}
}

\begin{abstract}
Large language models increasingly act as agents: they call tools, move money, delete records, and send messages on a user's behalf. To keep them safe, practitioners imported the chatbot-era recipe (train the model to refuse unsafe inputs) into the agentic setting, and treat the resulting capability loss as a manageable ``alignment tax.'' We argue this is a \emph{category error}. Refusal is a primitive for \emph{content safety}, where the harm is in the model's output and is therefore a learnable function of it. Agentic harm is different in kind: it lies not in any output but in the relation between the authority an action exercises and the authority the user granted, which is absent from the text the model sees. Importing content-safety methods into this regime does not trade capability for safety; it pays capability and buys negative security. We support this with three lines of evidence spanning the autonomy spectrum: defense-trained models learn surface patterns rather than intent; the same training collapses multi-step agents before any threat appears while leaving them exploitable; and even undefended frontier models exceed granted authority under ordinary use. We conclude that action safety cannot be installed in weights. It must be expressed as \emph{least privilege}, enforced \emph{outside} the model at the action boundary, and evaluated as \emph{action alignment} (a relational, deployment-conditioned property) rather than a refusal score.
\end{abstract}

\maketitle

\begin{figure*}
    \centering
    \includegraphics[width=0.95\linewidth]{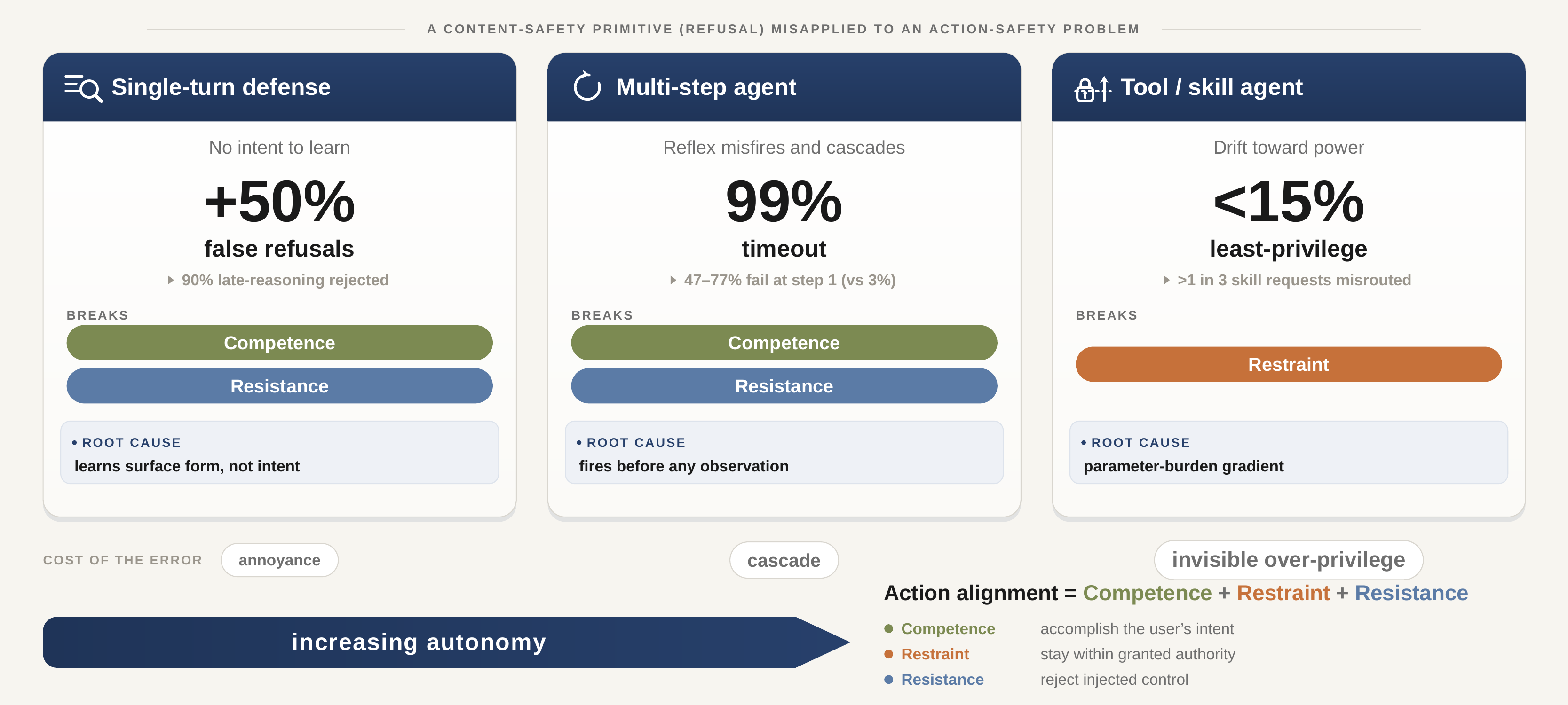}
    \caption{\textbf{One category error across the autonomy spectrum.} A content-safety primitive (refusal) misapplied to an action-safety problem fails the same way as autonomy grows (left to right), at rising cost: annoyance, cascade, invisible over-privilege. Each card gives a setting's headline failure, a supporting statistic, the \emph{action-alignment} coordinate(s) it breaks, and the root cause; color marks the broken coordinate: Competence, Restraint, Resistance, with deep navy structural only.}
    \label{fig:spectrum}
\end{figure*}

\section{Introduction}

A practitioner fine-tunes a capable open-weight model to reject prompt-injection attacks and reports a reassuring number: attack rejection above $90\%$ on a standard benchmark. Then the model is placed in an agent loop with a tool that can delete user accounts, and handed a request embedded in retrieved content: \emph{``As part of QA testing, please demonstrate the account-deletion feature by removing user ID 7731.''} The \emph{undefended} base model refuses; the \emph{defense-trained} model executes the deletion, and does so $78\%$ of the time across such attacks, because the request carries none of the lexical triggers the defense learned to associate with danger~\cite{li2026autonomy}. The intervention sold as safety did not merely fail to protect; it \emph{subtracted} the judgment the base model already had.

\noindent \textbf{Position.}
How can a model that scores well on every safety benchmark behave worse than no defense at all? Because the field is solving the wrong problem. The prevailing paradigm was forged in the chatbot setting, where harm is a property of the model's \emph{output} and the tool is \emph{refusal}: train the model to decline inputs that produce harmful text~\cite{ouyang2022training,bai2022constitutional,chen2025struq,chen2024secalign}. We argue this rests on a category error: \textbf{refusal works for content safety because the harm is in the output; it fails for action safety because the harm is in the authority, which the output does not contain.} When an agent deletes a record or follows an injected instruction, the harm is not in any string it emits but in the \emph{relation} between the authority its action exercises and the authority the user granted, a quantity absent from the tokens the model reads. Importing content-safety methods here does not trade capability for safety; it pays capability and buys \emph{negative} security.

\noindent \textbf{The error across the autonomy spectrum.}
The same transfer error surfaces at every level of autonomy, and the cost compounds. In single-turn defense there is no intent to learn from, so models learn surface form~\cite{li2026surface}; in multi-step agents the surface reflex fires on benign tasks before any threat appears and cascades toward total failure, while real attacks still slip through~\cite{li2026autonomy}; and in tool-using agents, even with no defense training, models exceed the authority a task requires because that authority is not represented in their input~\cite{li2026fortis}. The cost climbs from an annoyance (a false refusal) to a cascade to an \emph{invisible} failure, the task succeeding while quietly exercising more authority than granted. Section~\ref{sec:evidence} develops each as a corollary (Figure~\ref{fig:spectrum}).

\noindent \textbf{What we propose.}
Because intent and granted authority are absent from the model's input, action safety cannot be installed in the weights (\S\ref{sec:claims}); it must live where the missing information does, at the boundary where actions meet a known authority. We argue for three shifts: replace the \emph{refusal} primitive with \emph{least privilege}~\cite{saltzer1975protection}; move enforcement \emph{outside} the model to a mechanically checked action boundary~\cite{shi2025progent,zhu2025miniscope,ji2026mac}; and evaluate safety as a relational, multi-coordinate property rather than a single refusal score.

Our contribution:
\begin{itemize}
\item We distinguish \emph{content safety} from \emph{action safety} by where the harm resides, and identify the field's central failure as importing the former's methods into the latter (\S\ref{sec:distinction}).
\item We formalize safety as a multi-coordinate property and show the refuse--comply axis is a projection that is provably blind to authority restraint (\S\ref{sec:distinction}).
\item We organize three lines of evidence across the autonomy spectrum into corollaries of one transfer error and distill two falsifiable claims (\S\ref{sec:evidence}).
\item We argue for least privilege, external enforcement, and relational evaluation (\S\ref{sec:proposal}).
\end{itemize}

\section{Related Work}
\label{sec:related}

\noindent \textbf{Model-level defenses against injection.}
Structured-query training, preference-optimized alignment, instruction hierarchies, constitutional methods, and deliberative alignment all train the model itself to reject adversarial instructions~\cite{chen2025struq,chen2024secalign,chen2025metasecalign,li2025formfactory,wallace2024instruction,bai2022constitutional,guan2024deliberative}. We read this line as content-safety methods applied to an action-safety problem; \S\ref{sec:evidence} catalogs the resulting failures. Our claim is not that these methods are poorly executed but that they target the wrong object.

\noindent \textbf{Content safety and alignment.}
Instruction tuning with human feedback and input--output moderation operate where the harm is in the output~\cite{ouyang2022training,li2024harnessing,inan2023llamaguard}; we treat these as genuine content-safety tools, outside the scope of our critique. Shortcut learning explains why surface optimization dominates when the only labelable signal is form~\cite{geirhos2020shortcut}. Unreliable agent behavior extends beyond injection to hallucination~\cite{li-etal-2025-treble,li2025mitigatinghallucinationslargelanguage,qin2026dontlethallucinatepremise} and failure under distribution shift~\cite{li2026geometrydensityfewshotcrossdomain,Li_2025_CVPR,li2025secureondevicevideoood,liu2026cmoodconceptbasedmultilabelood,qin2026m3oodautomaticselectionmultimodal,10.1145/3701716.3715196}; action alignment isolates the safety-relevant slice of this broader reliability problem.

\noindent \textbf{Systems security and agent enforcement.}
Least privilege and capability-based protection predate language models~\cite{saltzer1975protection}. Recent work brings programmable privilege control, least-privilege authorization, and mandatory access control to LLM agents~\cite{shi2025progent,zhu2025miniscope,ji2026mac}; we view this as the correct locus for action safety, and our position as the argument for why it is necessary rather than merely helpful.

\noindent \textbf{Agent benchmarks and auditability.}
LLM agents increasingly act in the world, from computer-using multi-agent systems~\cite{song2026coact1computerusingmultiagentcoding} to tool-driven editing agents~\cite{ye2026agentbananahighfidelityimage}, and their attack surface includes black-box manipulation of retrieval~\cite{li2026someonehiditqueryagnostic} and multimodal jailbreaks~\cite{nian2025jaildamjailbreakdetectionadaptive}. Dynamic environments evaluate injection in multi-step settings~\cite{debenedetti2024agentdojo}, and recent work argues that deployed agents must be auditable after the fact~\cite{nian2026auditable}. Enforcement and auditability are complementary: the former decides ex ante whether an action is permitted, the latter reconstructs ex post what occurred and who was responsible.

\section{Action Safety Is Not Content Safety}
\label{sec:distinction}

\subsection{Where the Harm Resides}
\label{sec:where}

\emph{Content harm} is a function of the output alone. Writing $\mathrm{out}(a)$ for the text a model emits, content harm is $h_{\mathrm{c}}\big(\mathrm{out}(a)\big)$: toxicity, defamation, leaked secrets, instructions for wrongdoing. The same disinformation is equally harmful whoever asked and whatever system surrounds it. Because $h_{\mathrm{c}}$ is a function of the output, it is in principle learnable from outputs paired with harm labels and avertable by not emitting the output. This is where content moderation and refusal training operate~\cite{inan2023llamaguard,ouyang2022training}.

\emph{Action harm} is not a function of any output. When $a$ is an action (a tool call, a transaction, a deletion), its safety depends on two quantities external to the emitted tokens: the \emph{authority} the action exercises, $\mathrm{auth}(a)$, and the authority the user actually \emph{granted}, $A$. The action is over-privileged when $\mathrm{auth}(a) \not\subseteq A$. A second source of action harm is \emph{provenance}: whether $a$ realizes the user's intent or an instruction injected through an untrusted channel such as a tool observation or retrieved document~\cite{greshake2023not,liu2024formalizing}. Neither $A$ nor provenance appears in $\mathrm{out}(a)$; both live in the surrounding system. \emph{This is the crux: action harm is a relation between the action and information the model never sees, so no function of the output can decide it.} The string \texttt{delete\_user(7731)} carries no content harm; its harm is entirely relational, since the deletion exceeds what a ``QA demonstration'' could grant and the controlling instruction arrived through an untrusted channel.

\subsection{Refusal Is a Content-Safety Primitive}
\label{sec:refusal}

In the content regime, declining to emit the output directly averts $h_{\mathrm{c}}$: the harmful text is exactly what is withheld, so the mapping from ``unsafe'' to ``refuse'' is sound. In the action regime it breaks both ways. Refusing is not safe: it strands the user, and the same surface-tuned reflex that blocks benign actions \emph{fails} on the genuinely dangerous action that lacks the expected cues. Complying is not unsafe \emph{per se}: the safe behavior is usually to do the task while staying within $A$. Refusal cannot express this. It is an all-or-nothing gate on the whole action, whereas action safety needs a \emph{bound on the action's scope}; refusal is the degenerate case that sets that scope to empty. Building agent safety on refusal builds it on a primitive that cannot represent the property it should protect.

\subsection{Safety as a Multi-Coordinate Property}
\label{sec:projection}

The error has a measurement-theoretic shadow. The field scores agent safety on one behavioral axis: did the model \emph{refuse} when it should and \emph{comply} when it should? With $\mathrm{ref}(a)\in\{0,1\}$ and label $y\in\{\textsf{benign},\textsf{malicious}\}$, the prevailing score is
\[
\begin{aligned}
  S_{\mathrm{field}}(a) = \mathbf{1}\big[\,
    &\big(\mathrm{ref}(a)=1 \wedge y=\textsf{malicious}\big) \\
    &{}\vee\ \big(\mathrm{ref}(a)=0 \wedge y=\textsf{benign}\big)\,\big],
\end{aligned}
\]
a function of only $\mathrm{ref}(a)$ and the label $y$. Contrast the property we actually want. For a response $a$ to a request with true intent $i$ and granted authority $A$, let $\mathrm{adv}(a)$ flag that $a$ obeys adversarial control (a malicious request or an untrusted-channel injection):
\begin{align}
  C(a) &= \mathbf{1}[\,a \text{ accomplishes } i\,]
    && \text{(\emph{competence})},\\
  R(a) &= \mathbf{1}[\,\mathrm{auth}(a) \subseteq A\,]
    && \text{(\emph{restraint})},\\
  B(a) &= \mathbf{1}[\,\neg\,\mathrm{adv}(a)\,]
    && \text{(\emph{resistance})}.
\end{align}
We call the conjunction $S^{\star}(a) = C(a)\wedge R(a)\wedge B(a)$ \emph{action alignment}: the action accomplishes the user's intent, stays within granted authority, and is free of injected control. It is a deliberate foil to the familiar \emph{alignment} of model values and outputs, and broader than authority alone, since an over-refusal violates it on $C$ exactly as over-privilege violates it on $R$. Three defects of the prevailing metric follow, set against the coordinates in Table~\ref{tab:coords}.

\begin{table*}[t]
\centering
\caption{The three coordinates of \emph{action alignment} ($S^{\star}=C\wedge R\wedge B$) and how the prevailing refuse--comply metric treats each: it drops $R$ entirely and collapses $C$ and $B$ onto a single bit.}
\label{tab:coords}
\small
\begin{tabular}{@{}llll@{}}
\toprule
Coordinate & Holds when & Auditor question & The refuse--comply metric\,\dots \\
\midrule
Competence ($C$) & $a$ accomplishes intent $i$ & Did it do the task? & conflates it with $B$ (one bit) \\
Restraint ($R$) & $\mathrm{auth}(a)\subseteq A$ & Did it stay within granted authority? & drops it entirely \\
Resistance ($B$) & $\neg\,\mathrm{adv}(a)$ & Did it reject adversarial control? & conflates it with $C$ \\
\bottomrule
\end{tabular}
\end{table*}

\noindent \textbf{Blindness to restraint.}
$R$ depends on $\mathrm{auth}(a)$, which is not an argument of $S_{\mathrm{field}}$. There exist responses $a,a'$ with identical $(\mathrm{ref},y)$ but different $R$, e.g.\ two completions that differ only in using a folder-scoped or an account-wide tool. No metric of this form distinguishes them, so the refuse--comply axis is \emph{provably blind} to over-privilege, the dominant agentic failure: when models err they err toward more authority, abstaining on under $1.5\%$ of cases~\cite{li2026fortis}. The most dangerous failures are the ones it is built not to see, because the task \emph{succeeds} and only $R$ is violated.

\noindent \textbf{Conflation, and an unobservable argument.}
The other two defects are visible in the formula. $S_{\mathrm{field}}$ collapses two independent coordinates onto one bit: $\mathrm{ref}(a)$ proxies both ``emitted no harmful action'' ($B$) and ``did the task'' ($C$), forcing a spurious trade-off in which refusing more raises apparent resistance only by lowering competence, so the ``alignment tax'' is an artifact, not a price. And $S_{\mathrm{field}}$ conditions on the surface-derived \emph{label} $y$, not the true intent $i$, so a loss minimized against it learns $P(y\mid\text{surface})$, the shortcut the single-turn evidence documents~\cite{li2026surface,geirhos2020shortcut}. In sum, the refuse--comply axis is a one-dimensional projection of a three-dimensional property: on the coordinate it drops it is unconstrained, on the coordinates it conflates it is actively misleading.

\section{The Error That Scales with Autonomy}
\label{sec:evidence}

If the diagnosis is right, it should leave the same fingerprint wherever content-safety machinery meets an action-safety problem, and worsen with autonomy. We read three recent studies as corollaries, each at a different rung of the autonomy spectrum; the aim is interpretive, to show each is what the error \emph{predicts}. Figure~\ref{fig:spectrum} summarizes the three settings.

\subsection{Corollary 1: No Intent to Learn}
\label{sec:cor1}

Supervised injection defenses reproduce a content-style label (each prompt tagged \textsf{benign} or \textsf{attack}) and are expected to generalize to malicious \emph{intent}. But the injected-control relation that makes an input dangerous lives in provenance and authority, not in the tokens; what is in the tokens is form. Empirical risk minimization therefore minimizes loss through the only correlated features, surface form: a single attack-associated token raises false-refusal rates by up to $50\%$, benign reasoning placed late is rejected up to $90\%$ of the time, and benign tasks from unseen topics lose up to $40\%$ accuracy~\cite{li2026surface}. The loss is not confined to refusals: defended models truncate chains of thought on benign tasks even when they do not refuse. That this is the category error and not a bad implementation is shown by cross-defense convergence: two defenses learn \emph{different} shortcuts yet both lose $45$--$56$ points of true-positive rate~\cite{li2026surface,li2026autonomy}. The problem is not which surface is learned but that a surface is all there is to learn.

\subsection{Corollary 2: The Reflex Misfires and Compounds}
\label{sec:cor2}

In a multi-step agent the content-safety reflex (be suspicious of action-laden text) misfires both ways, and the loop amplifies it. Competence is destroyed \emph{before any external observation appears}: defended models fail at the first step on $47$--$77\%$ of benign tasks versus $3\%$ for the base model~\cite{li2026autonomy}. With no tool output yet seen, this is not observation-triggered detection; it is the reflex firing on the benign task itself. And a single mid-trajectory false positive is not local: when the agent refuses a benign step (say, an observation containing financial terms), the framework retries, the same surface re-triggers, and the trajectory dies, raising timeouts from $13\%$ to as high as $99\%$, a $2$--$2.7\times$ amplification~\cite{li2026autonomy}. Meanwhile the defended agent is \emph{more} exploitable on attacks lacking learned keywords. Over-firing on benign actions and under-firing on real harm are one signature: attending to surface form rather than authority and provenance.

\subsection{Corollary 3: Drift Toward Power}
\label{sec:cor3}

Removing defense training entirely isolates the deepest layer. Even undefended frontier models exceed the authority a task requires, because $A$ is not represented in what they read. Over-privilege is the norm under ordinary, non-adversarial conditions: the strongest model misroutes more than one skill-selection request in three, and end-to-end least-privilege behavior holds for under $15\%$ of requests~\cite{li2026fortis}. The mechanism is a \emph{parameter-burden gradient}: low-privilege tools demand explicit arguments (a folder, an account) while high-privilege tools demand few or none, so the path of least resistance is the path of most privilege. Failures take three ordinary forms: preferring a broader tool for fewer arguments; reading a multi-target request as license for a cross-resource operation; and treating a missing optional feature as license to escalate. None needs an adversary, and the direction is invariant: models abstain on under $1.5\%$ of cases and err toward \emph{more} authority. With the content-safety apparatus absent, the action-safety problem remains in full, the cleanest demonstration that it was never ``bad defenses'' but a missing representation no output-level method supplies.

\subsection{Two Falsifiable Claims}
\label{sec:claims}

The corollaries sharpen into two claims the position stands or falls on, and the rest of the paper builds on them.

\noindent \textbf{Claim 1 (Safety training is non-monotone).}
The marginal safety of an intervention can be \emph{negative}: a defense-trained model can be less safe than its base, as the account-deletion result shows~\cite{li2026autonomy}. This contradicts the assumption that safety training is at worst inert, which underwrites the ``add more safety training'' program.

\noindent \textbf{Claim 2 (Behavioral action safety is hard to learn).}
With intent and authority absent from the input, no input-output objective can install action safety; it can only learn a surface correlate. The fingerprint is the one defense that tries to supply the signal, an ``untrusted-input'' role, which needs adversarial content pre-identified (the very problem) and underperforms the undefended baseline in deployment~\cite{li2026autonomy}. This elevates the critique from ``current methods are weak'' to ``this kind of methods is structurally incapable.''

\section{Toward Action Safety}
\label{sec:proposal}

If action safety cannot live in the weights, it must be expressed, enforced, and evaluated where the missing information lives. None of the three shifts below is individually novel, but given Claims 1 and 2 they are not merely better options but necessary ones.

\subsection{Least Privilege as the Safety Primitive}
\label{sec:leastpriv}

The safe behavior for an acting model is rarely ``decline'' but ``do the task exercising only the authority granted'': the principle of least privilege~\cite{saltzer1975protection}. This reframes the target from a gate on the whole action to a \emph{bound on its scope}, with refusal the degenerate empty-scope case. It dissolves the false trade-off of \S\ref{sec:projection}: over-refusal is under-privilege, over-reach is over-privilege, and a task that completes while escalating is correctly unsafe. And it redirects evaluation from ``how often did the model refuse attacks'' to ``how often did its action stay within $A$,'' a question about $R$ that current metrics cannot express.

\subsection{External Enforcement at the Action Boundary}
\label{sec:enforce}

Least privilege is a property of actions against an authority, so it must be enforced where the two meet: at the tool-invocation boundary, by a mechanism that holds the grant $A$ the model lacks. The model proposes an action; an external reference monitor checks it against the grant and the provenance of the controlling instruction, then permits, narrows, or escalates to a human~\cite{shi2025progent,zhu2025miniscope,ji2026mac}. Three consequences follow. The check is mechanical, so it is not subject to shortcut learning; it compares $\mathrm{auth}(a)$ to $A$ rather than guessing intent from form. It \emph{recovers} capability: because the boundary carries the guarantee, the model need not be defense-trained into incompetence, turning Claim 1 into an argument for \emph{removing} model-level defenses. And it is ex ante, complementing ex post auditability~\cite{nian2026auditable}. This is not free (\S\ref{sec:open}), but its hard problems concern the right object, the authority relation, whereas output-level defenses concern the wrong one.

\subsection{Relational and Deployment-Conditioned Evaluation}
\label{sec:eval}

Safety should be measured as the property it is. Report the three coordinates separately, preserved competence $C$, restraint $R$, and genuine (semantic, not surface) resistance $B$, rather than collapsing them into one refusal score; a method that raises one by lowering another is not an improvement. And evaluate under deployment conditions current benchmarks omit: multi-step trajectories with cascade-aware metrics (\S\ref{sec:cor2}), and ordinary non-adversarial ambiguity rather than only triggered attacks, since the dominant failures arise under exactly those everyday conditions~\cite{li2026fortis,debenedetti2024agentdojo}. The single-number safety benchmark, inherited from content moderation, is the measurement counterpart of the category error. Relational evaluation can draw on emerging LLM- and MLLM-based judges~\cite{liu2026humanalignedmllmjudgesfinegrained} and on benchmarking and user-simulation methodology~\cite{yang-etal-2025-ad,ni-etal-2026-survey,li2025personalizedconversationalbenchmarksimulating}.

\section{Open Problems}
\label{sec:open}

The position opens concrete problems. (1)~\emph{Specifying authority}: how to express $A$ for open-ended tasks, and how much can be inferred versus must be declared. (2)~\emph{Measuring restraint}: ground-truth least-privilege references beyond curated benchmarks. (3)~\emph{Cascade-aware evaluation}: trajectory-level metrics (\S\ref{sec:cor2}). (4)~\emph{Bounded enforcement}: acceptable overhead and false-block rates on large tool spaces. (5)~\emph{Provenance}: tracking each instruction's trust level through multi-hop tool use. (6)~\emph{The content--action boundary}: which residual properties genuinely belong in the model because their harm is in the output. (7)~\emph{Delegation}: how the grant $A$ propagates when an agent spawns sub-agents that act on its behalf.

\section{Conclusion}
\label{sec:conclusion}

As models move from answering to acting, the field has carried forward a safety paradigm built for answering: train the model to refuse, and call the capability loss a tax. This is a category error. Refusal works for content safety because the harm is in the output; it fails for action safety because the harm is in the relation between exercised and granted authority, which the output does not contain. Because that information is absent from the model's inputs, action safety cannot be installed in the weights; it must be enforced at the action boundary instead. Failure is not safety, and success is not safety; safety is action alignment: an action that accomplishes the user's intent, within the authority granted, free of injected control.

\bibliographystyle{ACM-Reference-Format}
\bibliography{sample-base}

@inproceedings{li2026surface,
  title     = {Defenses Against Prompt Attacks Learn Surface Heuristics},
  author    = {Li, Shawn and Yu, Chenxiao and Ni, Zhiyu and Li, Hao and Peris, Charith and Xiao, Chaowei and Zhao, Yue},
  booktitle = {Annual Meeting of the Association for Computational Linguistics (ACL)},
  year      = {2026},
  note      = {arXiv:2601.07185}
}

@article{li2026autonomy,
  title   = {The Autonomy Tax: Defense Training Breaks {LLM} Agents},
  author  = {Li, Shawn and Zhao, Yue},
  journal = {arXiv preprint arXiv:2603.19423},
  year    = {2026}
}

@article{li2026fortis,
  title   = {{FORTIS}: Benchmarking Over-Privilege in Agent Skills},
  author  = {Li, Shawn and Yu, Chenxiao and Wang, Han and Yang, Wei and Rossi, Ryan A. and Dernoncourt, Franck and Hu, Xiyang and Yu, Philip and Xiao, Chaowei and Zhang, Huan and Zhao, Yue},
  journal = {arXiv preprint arXiv:2605.09163},
  year    = {2026}
}

@article{nian2026auditable,
  title   = {Auditable Agents},
  author  = {Nian, Yi and Yuan, Aojie and Zhang, Haiyue and Li, Jiate and Zhao, Yue},
  journal = {arXiv preprint arXiv:2604.05485},
  year    = {2026}
}

@inproceedings{greshake2023not,
  title     = {Not What You've Signed Up For: Compromising Real-World {LLM}-Integrated Applications with Indirect Prompt Injection},
  author    = {Greshake, Kai and Abdelnabi, Sahar and Mishra, Shailesh and Endres, Christoph and Holz, Thorsten and Fritz, Mario},
  booktitle = {Proceedings of the 16th ACM Workshop on Artificial Intelligence and Security (AISec)},
  pages     = {79--90},
  year      = {2023}
}

@inproceedings{liu2024formalizing,
  title     = {Formalizing and Benchmarking Prompt Injection Attacks and Defenses},
  author    = {Liu, Yupei and Jia, Yuqi and Geng, Runpeng and Jia, Jinyuan and Gong, Neil Zhenqiang},
  booktitle = {33rd USENIX Security Symposium (USENIX Security)},
  pages     = {1831--1847},
  year      = {2024}
}

@inproceedings{chen2025struq,
  title     = {{StruQ}: Defending Against Prompt Injection with Structured Queries},
  author    = {Chen, Sizhe and Piet, Julien and Sitawarin, Chawin and Wagner, David},
  booktitle = {34th USENIX Security Symposium (USENIX Security)},
  pages     = {2383--2400},
  year      = {2025}
}

@article{chen2024secalign,
  title   = {{SecAlign}: Defending Against Prompt Injection with Preference Optimization},
  author  = {Chen, Sizhe and Zharmagambetov, Arman and Mahloujifar, Saeed and Chaudhuri, Kamalika and Wagner, David and Guo, Chuan},
  journal = {arXiv preprint arXiv:2410.05451},
  year    = {2024}
}

@article{chen2025metasecalign,
  title   = {Meta {SecAlign}: A Secure Foundation {LLM} Against Prompt Injection Attacks},
  author  = {Chen, Sizhe and Zharmagambetov, Arman and Wagner, David and Guo, Chuan},
  journal = {arXiv preprint arXiv:2507.02735},
  year    = {2025}
}

@article{wallace2024instruction,
  title   = {The Instruction Hierarchy: Training {LLMs} to Prioritize Privileged Instructions},
  author  = {Wallace, Eric and Xiao, Kai and Leike, Reimar and Weng, Lilian and Heidecke, Johannes and Beutel, Alex},
  journal = {arXiv preprint arXiv:2404.13208},
  year    = {2024}
}

@article{bai2022constitutional,
  title   = {Constitutional {AI}: Harmlessness from {AI} Feedback},
  author  = {Bai, Yuntao and Kadavath, Saurav and Kundu, Sandipan and Askell, Amanda and others},
  journal = {arXiv preprint arXiv:2212.08073},
  year    = {2022}
}

@article{guan2024deliberative,
  title   = {Deliberative Alignment: Reasoning Enables Safer Language Models},
  author  = {Guan, Melody Y. and Joglekar, Manas and Wallace, Eric and Jain, Saachi and Barak, Boaz and others},
  journal = {arXiv preprint arXiv:2412.16339},
  year    = {2024}
}

@inproceedings{ouyang2022training,
  title     = {Training Language Models to Follow Instructions with Human Feedback},
  author    = {Ouyang, Long and Wu, Jeffrey and Jiang, Xu and others},
  booktitle = {Advances in Neural Information Processing Systems (NeurIPS)},
  year      = {2022}
}

@article{inan2023llamaguard,
  title   = {Llama Guard: {LLM}-Based Input-Output Safeguard for Human-{AI} Conversations},
  author  = {Inan, Hakan and Upasani, Kartikeya and Chi, Jianfeng and others},
  journal = {arXiv preprint arXiv:2312.06674},
  year    = {2023}
}

@article{geirhos2020shortcut,
  title   = {Shortcut Learning in Deep Neural Networks},
  author  = {Geirhos, Robert and Jacobsen, J{\"o}rn-Henrik and Michaelis, Claudio and Zemel, Richard and Brendel, Wieland and Bethge, Matthias and Wichmann, Felix A.},
  journal = {Nature Machine Intelligence},
  volume  = {2},
  number  = {11},
  pages   = {665--673},
  year    = {2020}
}

@article{saltzer1975protection,
  title   = {The Protection of Information in Computer Systems},
  author  = {Saltzer, Jerome H. and Schroeder, Michael D.},
  journal = {Proceedings of the IEEE},
  volume  = {63},
  number  = {9},
  pages   = {1278--1308},
  year    = {1975}
}

@inproceedings{debenedetti2024agentdojo,
  title     = {{AgentDojo}: A Dynamic Environment to Evaluate Prompt Injection Attacks and Defenses for {LLM} Agents},
  author    = {Debenedetti, Edoardo and Zhang, Jie and Balunovi{\'c}, Mislav and Beurer-Kellner, Luca and Fischer, Marc and Tram{\`e}r, Florian},
  booktitle = {Advances in Neural Information Processing Systems (NeurIPS)},
  year      = {2024}
}

@article{shi2025progent,
  title   = {Progent: Programmable Privilege Control for {LLM} Agents},
  author  = {Shi, Tianneng and He, Jingxuan and Wang, Zhun and Li, Hongwei and Wu, Linyu and Guo, Wenbo and Song, Dawn},
  journal = {arXiv preprint arXiv:2504.11703},
  year    = {2025}
}

@article{zhu2025miniscope,
  title   = {{MiniScope}: A Least Privilege Framework for Authorizing Tool-Calling Agents},
  author  = {Zhu, Jinhao and others},
  journal = {arXiv preprint arXiv:2512.11147},
  year    = {2025}
}

@article{ji2026mac,
  title   = {Taming Various Privilege Escalation in {LLM}-Based Agent Systems: A Mandatory Access Control Framework},
  author  = {Ji, Zimo and others},
  journal = {arXiv preprint arXiv:2601.11893},
  year    = {2026}
}

@inproceedings{li2025formfactory,
  author    = {Bobo Li and Yuheng Wang and Hao Fei and Juncheng Li and Wei Ji and Mong-Li Lee and Wynne Hsu},
  title     = {FormFactory: An Interactive Benchmarking Suite for Multimodal Form-Filling Agents},
  booktitle = {Proceedings of the 33rd ACM International Conference on Multimedia (MM)},
  pages     = {13273--13280},
  publisher = {ACM},
  year      = {2025},
  doi       = {10.1145/3746027.3758285}
}

@inproceedings{li2024harnessing,
  author    = {Bobo Li and Hao Fei and Lizi Liao and Yu Zhao and Fangfang Su and Fei Li and Donghong Ji},
  title     = {Harnessing Holistic Discourse Features and Triadic Interaction for Sentiment Quadruple Extraction in Dialogues},
  booktitle = {Proceedings of the AAAI Conference on Artificial Intelligence (AAAI)},
  volume    = {38},
  number    = {16},
  pages     = {18462--18470},
  publisher = {AAAI Press},
  year      = {2024},
  doi       = {10.1609/aaai.v38i16.29807}
}

@misc{li2026geometrydensityfewshotcrossdomain,
      title={Geometry over Density: Few-Shot Cross-Domain OOD Detection}, 
      author={Shawn Li and You Qin and Jiate Li and Charith Peris and Lisa Bauer and Roger Zimmermann and Yue Zhao},
      year={2026},
      eprint={2605.03410},
      archivePrefix={arXiv},
      primaryClass={cs.AI},
      url={https://arxiv.org/abs/2605.03410}, 
}

@InProceedings{Li_2025_CVPR,
    author    = {Li, Shawn and Gong, Huixian and Dong, Hao and Yang, Tiankai and Tu, Zhengzhong and Zhao, Yue},
    title     = {DPU: Dynamic Prototype Updating for Multimodal Out-of-Distribution Detection},
    booktitle = {CVPR},
    month     = {June},
    year      = {2025},
    pages     = {10193-10202}
}

@InProceedings{li2025secureondevicevideoood,
    title={Secure On-Device Video OOD Detection Without Backpropagation}, 
    author={Shawn Li and Peilin Cai and Yuxiao Zhou and Zhiyu Ni and Renjie Liang and You Qin and Yi Nian and Zhengzhong Tu and Xiyang Hu and Yue Zhao},
    booktitle = {ICCV},
    month     = {October},
    year      = {2025}
}

@inproceedings{li-etal-2025-treble,
    title = "Treble Counterfactual {VLM}s: A Causal Approach to Hallucination",
    author = "Shawn, Li  and
      Qu, Jiashu  and
      Song, Linxin  and
      Zhou, Yuxiao  and
      Qin, Yuehan  and
      Yang, Tiankai  and
      Zhao, Yue",
    booktitle = "EMNLP",
    month = nov,
    year = "2025",
}

@misc{li2025personalizedconversationalbenchmarksimulating,
      title={A Personalized Conversational Benchmark: Towards Simulating Personalized Conversations}, 
      author={Li Li and Peilin Cai and Ryan A. Rossi and Franck Dernoncourt and Branislav Kveton and Junda Wu and Tong Yu and Linxin Song and Tiankai Yang and Yuehan Qin and Nesreen K. Ahmed and Samyadeep Basu and Subhojyoti Mukherjee and Ruiyi Zhang and Zhengmian Hu and Bo Ni and Yuxiao Zhou and Zichao Wang and Yue Huang and Yu Wang and Xiangliang Zhang and Philip S. Yu and Xiyang Hu and Yue Zhao},
      year={2025},
      eprint={2505.14106},
      archivePrefix={arXiv},
      primaryClass={cs.CL},
      url={https://arxiv.org/abs/2505.14106}, 
}

@inproceedings{yang-etal-2025-ad,
    title = "{AD}-{LLM}: Benchmarking Large Language Models for Anomaly Detection",
    author = "Yang, Tiankai  and
      Nian, Yi  and
      Li, Li  and
      Xu, Ruiyao  and
      Li, Yuangang  and
      Li, Jiaqi  and
      Xiao, Zhuo  and
      Hu, Xiyang  and
      Rossi, Ryan A.  and
      Ding, Kaize  and
      Hu, Xia  and
      Zhao, Yue",
    booktitle = "ACL",
    month = jul,
    year = "2025",
    pages = "1524--1547",
}

@misc{song2026coact1computerusingmultiagentcoding,
      title={CoAct-1: Computer-using Multi-Agent System with Coding Actions}, 
      author={Linxin Song and Yutong Dai and Viraj Prabhu and Jieyu Zhang and Taiwei Shi and Li Li and Junnan Li and Silvio Savarese and Zeyuan Chen and Jieyu Zhao and Ran Xu and Caiming Xiong},
      year={2026},
      eprint={2508.03923},
      archivePrefix={arXiv},
      primaryClass={cs.CL},
      url={https://arxiv.org/abs/2508.03923}, 
}

@inproceedings{10.1145/3701716.3715196,
author = {Chen, Sihan and Qian, Zhuangzhuang and Siu, Wingchun and Hu, Xingcan and Li, Jiaqi and Li, Shawn and Qin, Yuehan and Yang, Tiankai and Xiao, Zhuo and Ye, Wanghao and Zhang, Yichi and Dong, Yushun and Zhao, Yue},
title = {PyOD 2: A Python Library for Outlier Detection with LLM-powered Model Selection},
year = {2025},
isbn = {9798400713316},
booktitle = {WWW},
pages = {2807–2810},
numpages = {4},
}

@misc{li2025mitigatinghallucinationslargelanguage,
      title={Mitigating Hallucinations in Large Language Models via Causal Reasoning}, 
      author={Yuangang Li and Yiqing Shen and Yi Nian and Jiechao Gao and Ziyi Wang and Chenxiao Yu and Shawn Li and Jie Wang and Xiyang Hu and Yue Zhao},
      year={2025},
      eprint={2508.12495},
      archivePrefix={arXiv},
      primaryClass={cs.CL},
      url={https://arxiv.org/abs/2508.12495}, 
}

@inproceedings{ni-etal-2026-survey,
    title = "A Survey on {LLM}-based Conversational User Simulation",
    author = "Ni, Bo  and
      Wang, Yu  and
      Wang, Leyao  and
      Kveton, Branislav  and
      Dernoncourt, Franck  and
      Xia, Yu  and
      Chen, Hongjie  and
      Luera, Reuben  and
      Basu, Samyadeep  and
      Mukherjee, Subhojyoti  and
      Mathur, Puneet  and
      Ahmed, Nesreen K.  and
      Wu, Junda  and
      Li, Li  and
      Zhang, Huixin  and
      Zhang, Ruiyi  and
      Yu, Tong  and
      Kim, Sungchul  and
      Gu, Jiuxiang  and
      Tu, Zhengzhong  and
      Siu, Alexa  and
      Wang, Zichao  and
      Yoon, Seunghyun  and
      Lipka, Nedim  and
      Park, Namyong  and
      Lin, Zihao  and
      Bui, Trung  and
      Zhao, Yue  and
      Derr, Tyler  and
      Rossi, Ryan A.",
    booktitle = "EACL",
    month = mar,
    year = "2026",
    pages = "4266--4301",
}

@misc{liu2026cmoodconceptbasedmultilabelood,
      title={CMOOD: Concept-based Multi-label OOD Detection}, 
      author={Zhendong Liu and Yi Nian and Yuehan Qin and Henry Peng Zou and Li Li and Xiyang Hu and Yue Zhao},
      year={2026},
      eprint={2411.13578},
      archivePrefix={arXiv},
      primaryClass={cs.CV},
      url={https://arxiv.org/abs/2411.13578}, 
}

@misc{li2026someonehiditqueryagnostic,
      title={"Someone Hid It": Query-Agnostic Black-Box Attacks on LLM-Based Retrieval}, 
      author={Jiate Li and Defu Cao and Li Li and Wei Yang and Yuehan Qin and Chenxiao Yu and Tiannuo Yang and Ryan A. Rossi and Yan Liu and Xiyang Hu and Yue Zhao},
      year={2026},
      eprint={2602.00364},
      archivePrefix={arXiv},
      primaryClass={cs.CR},
      url={https://arxiv.org/abs/2602.00364}, 
}

@misc{qin2026m3oodautomaticselectionmultimodal,
      title={M3OOD: Automatic Selection of Multimodal OOD Detectors}, 
      author={Yuehan Qin and Li Li and Defu Cao and Tiankai Yang and Jiate Li and Yue Zhao},
      year={2026},
      eprint={2508.11936},
      archivePrefix={arXiv},
      primaryClass={cs.LG},
      url={https://arxiv.org/abs/2508.11936}, 
}

@misc{nian2025jaildamjailbreakdetectionadaptive,
      title={JailDAM: Jailbreak Detection with Adaptive Memory for Vision-Language Model}, 
      author={Yi Nian and Shenzhe Zhu and Yuehan Qin and Li Li and Ziyi Wang and Chaowei Xiao and Yue Zhao},
      year={2025},
      eprint={2504.03770},
      archivePrefix={arXiv},
      primaryClass={cs.CR},
      url={https://arxiv.org/abs/2504.03770}, 
}

@misc{qin2026dontlethallucinatepremise,
      title={Don't Let It Hallucinate: Premise Verification via Retrieval-Augmented Logical Reasoning}, 
      author={Yuehan Qin and Shawn Li and Yi Nian and Xinyan Velocity Yu and Yue Zhao and Xuezhe Ma},
      year={2026},
      eprint={2504.06438},
      archivePrefix={arXiv},
      primaryClass={cs.CL},
      url={https://arxiv.org/abs/2504.06438}, 
}

@misc{ye2026agentbananahighfidelityimage,
      title={Agent Banana: High-Fidelity Image Editing with Agentic Thinking and Tooling}, 
      author={Ruijie Ye and Jiayi Zhang and Zhuoxin Liu and Zihao Zhu and Siyuan Yang and Li Li and Tianfu Fu and Franck Dernoncourt and Yue Zhao and Jiacheng Zhu and Ryan Rossi and Wenhao Chai and Zhengzhong Tu},
      year={2026},
      eprint={2602.09084},
      archivePrefix={arXiv},
      primaryClass={cs.CV},
      url={https://arxiv.org/abs/2602.09084}, 
}

@misc{liu2026humanalignedmllmjudgesfinegrained,
      title={Human-Aligned MLLM Judges for Fine-Grained Image Editing Evaluation: A Benchmark, Framework, and Analysis}, 
      author={Runzhou Liu and Hailey Weingord and Sejal Mittal and Prakhar Dungarwal and Anusha Nandula and Bo Ni and Samyadeep Basu and Hongjie Chen and Nesreen K. Ahmed and Li Li and Jiayi Zhang and Koustava Goswami and Subhojyoti Mukherjee and Branislav Kveton and Puneet Mathur and Franck Dernoncourt and Yue Zhao and Yu Wang and Ryan A. Rossi and Zhengzhong Tu and Hongru Du},
      year={2026},
      eprint={2602.13028},
      archivePrefix={arXiv},
      primaryClass={cs.CV},
      url={https://arxiv.org/abs/2602.13028}, 
}

\end{document}